\documentclass{article}

\usepackage[final, nonatbib]{new_in_ML}

\usepackage[utf8]{inputenc} 
\usepackage[T1]{fontenc}    
\usepackage{hyperref}       
\usepackage{url}            
\usepackage{booktabs}       
\usepackage{amsfonts}       
\usepackage{amsmath}
\usepackage{nicefrac}       
\usepackage{microtype}      
\usepackage{xcolor}         
\newtheorem{theorem}{Theorem}[section]
\usepackage{graphicx}
\usepackage{subfigure}  

\title{OpenTensor: Reproducing Faster Matrix Multiplication Discovering Algorithms}

\author{Yiwen Sun$^{1}$\thanks{These authors contributed equally to this work.} \quad Wenye Li$^{1*}$ \\
$^1$School of Data Science, Fudan University, Shanghai, China. \\
\texttt{\{Email:ywsun22@m.fudan.edu.cn\}}}

\begin{document}

\maketitle

\begin{abstract}
OpenTensor is a reproduction of AlphaTensor, which discovered a new algorithm that outperforms the state-of-the-art methods for matrix multiplication by Deep Reinforcement Learning (DRL). While AlphaTensor provides a promising framework for solving scientific problems,  it is really hard to reproduce due to the massive tricks and lack of source codes. In this paper, we clean up the algorithm pipeline, clarify the technical details, and make some improvements to the training process. Computational results show that OpenTensor can successfully find efficient matrix multiplication algorithms.
\end{abstract}

\section{Introduction}
Matrix multiplication (MM) is a fundamental numerical operation that is used everywhere. To search for faster MM algorithms, DeepMind proposed AlphaTensor \cite{alphatensor} based on AlphaZero \cite{alphazero} and constructed a Monte Carlo Tree Search (MCTS) architecture. AlphaTensor \cite{alphatensor} not only finds a faster algorithm for matrix multiplication but also provides a new paradigm for using machine learning to solve scientific problems. However, due to the lack of open-source codes and too many algorithmic tricks, researchers may get lost in the myriad of details and find it hard to understand the key points, let alone reproduce the performance and implement it to solve other problems.

In this paper, we reproduce AlphaTensor \cite{alphatensor} and hope that it will be helpful for others to fully understand the scientific problem-solving paradigm. To distinguish  from the official algorithm, we coin the reproduction OpenTensor, which includes almost all tricks in the original work, such as synthetic demonstrations, change of basis, order shuffling, and action canonicalization. 

It is well known that matrix multiplication is indeed a tensor decomposition problem \cite{tensor}, which can be further reformulated as a search problem. Thus we can develop a  computational method by combining deep neural networks and MCTS. In OpenTensor, synthetic demonstrations are generated for training the neural network in a way of supervised learning. Then, we use MCTS to decompose tensors and canonicalize actions to avoid ambiguity. Tricks in the original work \cite{alphatensor} including basis transformation and order shuffling are also performed for data augmentation.

What is more, we slightly change the way of generating synthetic demonstrations in the original algorithm \cite{alphatensor} to alleviate the overestimation of tensor rank. We also  make a change of swapping actions during the training process to avoid bottlenecks for data preparation. Those changes can help reduce the bias in synthetic data and make OpenTensor converge faster
\footnote{Our codes are available on https://github.com/YiwenAI/OpenTensor.}.

\section{Algorithm and Training Details}

\subsection{Problem Formulation and Algorithm Pipeline}
\paragraph{Problem Formulation.} As already mentioned, matrix multiplication corresponds to a tensor decomposition problem. In particular, given a three-dimensional tensor $T$ of $S \times S \times S$, the goal here is to find a decomposition that is as short as possible. This can be formally expressed as:
\begin{equation}\label{eq:tensordecomp}
\mathrm{min} \enspace r 
\quad \mathrm{s.t.} \enspace T = \sum_{i=1}^{r} u_i \otimes v_i \otimes w_i,
\end{equation}
where the optimized variables are $\{(u_i, v_i, w_i)\}_{i=1}^{r}$. For the matrix multiplication problem, $T$ is the matrix multiplication tensor and factors $\{(u_i, v_i, w_i)\}_{i=1}^{r}$ can be used to construct an matrix 
product algorithm with $r$ multiplications\cite{tensor}.

\paragraph{Algorithm.} We can further reformulate  problem \eqref{eq:tensordecomp} as a search problem and solve it recursively. OpenTensor searches for one factor $(u_i, v_i, w_i)$ each time and the objective tensor is updated as follows until it becomes a zero tensor:
\begin{align}
\label{2}
    T_{t} \leftarrow T_{t-1} - (u_t \otimes v_t \otimes w_t).
\end{align}

Following AlphaTensor \cite{alphatensor}, we combine deep network and MCTS to select each factor. For the target tensor $T_t$, the neural network $f_\theta$ outputs {the factor selecting policy} $\pi \in \mathbb{R}^{|\mathcal{A}|}$ and {rank estimation} $v\in \mathbb{R}$.
Policy $\pi$ provides a proposal on factors space, and one can sample using $\pi$ by splitting factors into pieces and inferring autoregressively. After sampling using $\pi$ to get several factors candidates, we use MCTS with the guide of rank estimation $v$ to select one of them as the final factor. MCTS expands the current tensor $T_t$ with the candidate factors, selects factors to move forward, and evaluates new tensors with rank estimation by the neural network. After sufficient expansions, MCTS finds the best factor which is on the path to the tensor with the lowest rank:
\begin{align}
    (u_t, v_t, w_t) \leftarrow \mathrm{MCTS}(T_t\, | \, f_\theta).
\end{align}

\paragraph{Training.} Due to the lack of computing resources,  OpenTensor is mainly trained under the supervision of synthetic data.  We follow AlphaTensor \cite{alphatensor} to synthesize tensors by randomly generating factors and compositing them. Note that once the tensor is synthesized, we get one of its decompositions. OpenTensor's network is then optimized by these decomposition demonstrations in the way of supervised learning. Tricks of AlphaTensor training\cite{alphatensor}, including basis change, action canonicalization and shuffling, are performed in OpenTensor as well. 

However, the network trained by the synthetic data above always overestimates the rank of tensors, since the tensor may have an unknown decomposition of fewer factors than the way it is synthesized. To reduce the potential redundancy in the tensor synthesis, we identify a necessary condition for the optimal decomposition to filter the training data. Such improvement helps OpenTensor converge faster in the training process.

\subsection{Training Details}
\paragraph{Change of Basis} The rank of a tensor has the following property: changing the basis of a tensor does not change its rank. If a tensor $T$ can be decomposed as $T = \sum_{i=1}^{r} u_i \otimes v_i \otimes w_i$, then 
\begin{equation*}
T^{(A, B, C)} = \sum_{i=1}^{r} Au_i \otimes Bv_i \otimes Cw_i
\end{equation*}
defines a decomposition for $T^{(A, B, C)}$, where $A, B, C$ are invertible matrices. It follows that, if $\sum_{i=1}^{r} u_i \otimes v_i \otimes w_i$ is the best decomposition of $T$, then $\sum_{i=1}^{r} Au_i \otimes Bv_i \otimes Cw_i$ is also the best decomposition of $T^{(A,B,C)}$, and vise versa. Hence we follow AlphaTensor \cite{alphatensor} to randomly generate transformation matrices and apply basis transformation to the synthetic tensors as an augmentation.

\paragraph{Action Canonicalization}
Note that for any $\lambda_1 \lambda_2 \lambda_3=1$, $\lambda_1, \lambda_2, \lambda_3 \in \{1, -1\}$, factor $(u, v, w)$ is equivalent to $(\lambda_1u, \lambda_2v, \lambda_3w)$ after multiplication. In order to eliminate such redundancy, we assume that the first non-zero element in $u, v$ are always positive, and such actions are represented as standardized actions. If $f_{\theta}$ selects a non-standardized action, it will be regarded as sampling the corresponding standardized action. That is, we only retain the standard action in the end.

\paragraph{Order Shuffling}As factorizations are 
order invariant, we build an additional tensor-factorization training pair by swapping a random action with other actions from each finished game. We shuffle the order of actions in training dataset every ten epochs and reorder them before training.

\paragraph{Synthetic Demonstrations} In AlphaTensor \cite{alphatensor}, synthetic demonstrations are generated for pretraining to accelerate the decomposition process. However, we find that original generation method always overestimates the rank, since the number of generated factors is used as the groundtruth of rank while the optimal decomposition might contain fewer factors. To improve it, we have identified a necessary condition for checking whether a decomposition of the tensor has redundancy:


\begin{theorem}
    [Necessary Condition of No Redundancy] Assume $\mathcal{M}$ is the full set of $m \in \mathbb{R}^S$ used to generate tensors:
    \begin{align}
        T = \sum_{i=1}^{R} m_{ui} \otimes m_{vi} \otimes m_{wi}, \quad m_{ui}, m_{vi}, m_{wi} \in \mathcal{M},
    \end{align}
    and let the index set be $I = \{ u,v,w\}.$ For an arbitrary $i$, if the collection
    \begin{align}
        M_i = \left\{ m_{xi} \cdot m_{yi}^\top \; | \; x, y \in I, \;\; x \neq y \right\}
    \end{align}
    contains the same elements, then the decomposition of $\;T$ has redundancy.
\end{theorem}

We can apply this theorem to filter the synthetic data and it is found that over 70\% of original synthesized tensors contain redundancy. 

\section{Results and Summary}
\begin{figure}
  \centering
  \subfigure[Our Matrix Multiplication Algorithm]{
     \includegraphics[width=0.48\textwidth]{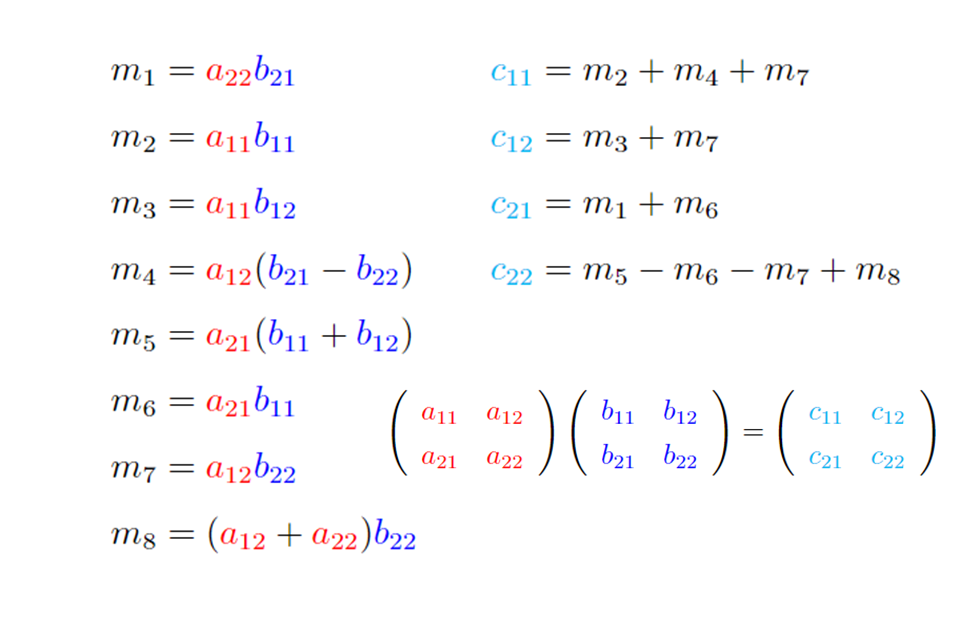}
  }
  \subfigure[Loss]{
     \includegraphics[width=0.40\textwidth]{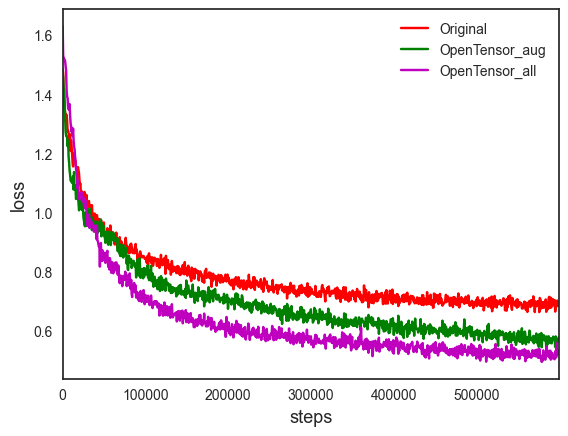}
  }
  \caption{Results of OpenTensor and Decomposition Solution. Left is the matrix multiplication algorithm found by OpenTensor ($2 \times 2$ matrix product with 8 multiplications), and right is the loss of three different methods. The original algorithm (\textcolor{red}{red}) contains action canonicalization, change of basis and original synthetic demonstrations technique. OpenTensor with augmentation (\textcolor{green}{green}) adds order shuffling and the final OpenTensor (\textcolor{violet}{purple}) uses all techniques we have mentioned. }
\label{result}

\end{figure}

We evaluate OpenTensor on RTX 4090, and each experiment runs about 40 GPU hours. The result is shown in Figure \ref{result}, which shows that our improvement helps the model converge better and faster. We also find that OpenTensor is sensitive to hyperparameters, while hyperparameters from the original paper are almost the best. 

In summary, OpenTensor faithfully replicates the implementation process of AlphaTensor and makes improvements on synthetic demonstrations. OpenTensor can efficiently decompose tensors and discover the matrix multiplication algorithm.

\bibliographystyle{plain}
\bibliography{main}


\end{document}